\documentclass[pdflatex,sn-chicago]{sn-jnl}

\jyear{2022}

\theoremstyle{thmstyleone}

\theoremstyle{thmstyletwo}

\theoremstyle{thmstylethree}

\begin{document}

\title[SETAR-Tree: A Novel and Accurate Tree Algorithm]{SETAR-Tree: A Novel and Accurate Tree Algorithm for Global Time Series Forecasting}

\author*[1]{\fnm{Rakshitha} \sur{Godahewa}}\email{rakshitha.godahewa@monash.edu}

\author[1]{\fnm{Geoffrey I.\ } \sur{Webb}}\email{geoff.webb@monash.edu}

\author[1]{\fnm{Daniel} \sur{Schmidt}}\email{daniel.schmidt@monash.edu}

\author[1]{\fnm{Christoph} \sur{Bergmeir}}\email{christoph.bergmeir@monash.edu}

\affil[1]{\orgdiv{Department of Data Science and AI}, \orgname{Monash University}, \orgaddress{\city{Melbourne}, \state{Victoria}, \country{Australia}}}

\abstract{Threshold Autoregressive (TAR) models have been widely used by statisticians for non-linear time series forecasting during the past few decades, due to their simplicity and mathematical properties. On the other hand, in the forecasting community, general-purpose tree-based regression algorithms (forests, gradient-boosting) have become popular recently due to their ease of use and accuracy. In this paper, we explore the close connections between TAR models and regression trees. These enable us to use the rich methodology from the literature on TAR models to define a hierarchical TAR model as a regression tree that trains globally across series, which we call SETAR-Tree. In contrast to the general-purpose tree-based models that do not primarily focus on forecasting, and calculate averages at the leaf nodes, we introduce a new forecasting-specific tree algorithm that trains global Pooled Regression (PR) models in the leaves allowing the models to learn cross-series information and also uses some time-series-specific splitting and stopping procedures. The depth of the tree is controlled by conducting a statistical linearity test commonly employed in TAR models, as well as measuring the error reduction percentage at each node split. Thus, the proposed tree model requires minimal external hyperparameter tuning and provides competitive results under its default configuration. We also use this tree algorithm to develop a forest where the forecasts provided by a collection of diverse SETAR-Trees are combined during the forecasting process. In our evaluation on eight publicly available datasets, the proposed tree and forest models are able to achieve significantly higher accuracy than a set of state-of-the-art tree-based algorithms and forecasting benchmarks across four evaluation metrics.}


\maketitle

\section{Introduction}
\label{sec:introduction}

Global time series forecasting models \citep[GFM,][]{ref_105} have shown their capability of providing accurate forecasts compared to traditional univariate forecasting models such as Exponential Smoothing~\citep[ETS,][]{ref_112} and Auto-Regressive Integrated Moving Average models~\citep[ARIMA,][]{ref_113}, in particular with obtaining superior results in the recently held M4 \citep{ref_30} and M5 \citep{makridakis_2020_m5} forecasting competitions. In contrast to traditional univariate forecasting models that build isolated models to forecast each series, GFMs are trained across collections of time series allowing the models to learn cross-series information, which also allows them to generalise better and control model complexity on a global level \citep{pablo_2020_principles}.

While a lot of research in this space focuses on neural networks and deep-learning models \citep[e.g.,][]{ref_99,ref_6,oreshkin_2019_nbeats}, tree-based GFM algorithms have risen greatly in popularity after dominating the winning methods of the M5 forecasting competition \citep{makridakis_2020_m5}. In particular, these approaches use \textit{LightGBM} \citep{guolin_lightgbm_2017} which is a highly efficient gradient boosted tree-based algorithm, to train global models across time series. Also other tree-based algorithms that can be trained as GFMs such as eXtreme Gradient Boosting \citep[XGBoost,][]{chen_xgboost_2016} and CatBoost \citep{NEURIPS2018_14491b75} have been used in the forecasting space. 
The advantages of these methods (especially compared with deep-learning methods) are their speed of training, their ease to include external regressors (numerical and categorical), and their relative ease and robustness of hyperparameter tuning. However, these algorithms are general-purpose algorithms, and none of them is specifically designed for forecasting. When performing regression, they use the average of the training outputs at a leaf node as the prediction for an instance that reaches that node during forecasting. Furthermore, while potentially easier to handle than in other algorithms, their accuracy is still subject to tuning of a number of hyperparameters such as the maximum tree depth, maximum number of leaf nodes, minimum number of instances in a leaf node and learning rate.
These findings motivate us to develop a global tree-based algorithm specifically for forecasting, achieving higher forecasting accuracy with minimal hyperparameter tuning. 

Piecewise linear models such as Threshold Autoregressive (TAR) models \citep{tong_1977_threshold,tong_1993_setar,Terasvirta_1994_star} have greatly contributed to the econometric and forecasting fields during the past few decades. 
These models partition the instance space of a particular single time series to be forecast into several subspaces where each subspace is modeled by a separate Autoregressive (AR) model. There are many variations of TAR models such as Self Exciting Threshold Autoregressive \citep[SETAR,][]{tong_1993_setar}, Smooth Transition Autoregressive \citep[STAR,][]{Terasvirta_1994_star}, Exponential STAR (ESTAR), Logistic STAR (LSTAR) and Neuro-Coefficient STAR \citep{medeiros_2000_hybrid} models.

Similar to the works of \citet{Aznarte_5556012}, \citet{aznarte_00506473} and \citet{AZNARTEM20072734} that establish analogies between certain types of TAR models and Fuzzy Rule Based Systems, we exploit the analogy of TAR models and decision trees that fit a linear model per leaf, for global time series forecasting, to introduce a new tree-based algorithm that uses the underlying concept of SETAR models in defining the splits. Thus, we name this model as \textit{SETAR-Tree}. In particular, our method internally finds the optimal past lag and the threshold that should be used to split the instances of a particular tree node into child nodes. The instances whose corresponding values of the optimal lag are greater than or equal to the selected optimal threshold and less than the selected optimal threshold are grouped separately whereas each instance group is considered as a child node. In contrast to the traditional general purpose tree-based regression algorithms that compute the average of the training outputs at a leaf node as the final prediction, our proposed method is a forecasting-specific tree which builds a Pooled Regression model \citep[PR,][]{gelman_hill_2006, pablo_2020_principles} which is a global linear AR model, at each leaf node allowing the model to learn cross-series information. As our model is a forecasting-specific tree with global regression models in the leaf nodes, it can be essentially identified as a global (cross-series) hierarchical SETAR model. Furthermore, the SETAR-Tree uses some time-series-specific stopping procedures. Exploiting the analogy of TAR models and regression trees, our proposed SETAR-Tree internally tunes the maximum tree depth using a statistical linearity test well established in the literature for TAR models \citep{Terasvirta_1994_star}, as well as measuring the error reduction percentage at each node split. Thus, our proposed model requires minimal external hyperparameter tuning whereas it provides competitive results even with its default hyperparameters. The proposed method is also applicable to time series forecasting problems that use external covariates during modelling.

Proposing a novel tree algorithm for forecasting naturally lends itself to extend the procedure to a forest, which is another main contribution of this study. In line with the traditional forest-based algorithms such as Random Forest \citep[RF,][]{breiman_2001_random}, our forest model uses a collection of diverse SETAR-Trees that are trained using a set of randomly selected time series. The forecasts provided by all trees are averaged to obtain the final forecasts. The trees are made diverse in terms of the significance of the statistical linearity test and error reduction percentage that are used to split each node. Thus, the forecasting accuracy of the forest model is even less sensitive to the hyperparameters than the tree.

Our proposed SETAR-Tree and SETAR-Forest algorithms in particular also outperform a set of state-of-the-art tree-based algorithms and forecasting benchmarks with statistical significance across eight experimental datasets on four error metrics. All implementations of this study are publicly available at:
\url{https://github.com/rakshitha123/SETAR_Trees}.

The remainder of this paper is organized as follows: Section \ref{sec:related_work} reviews the relevant prior work. Section \ref{sec:methodology} explains the theoretical concepts of SETAR models, and the proposed tree and forest algorithms in detail. Section \ref{sec:experiments} explains the experimental framework, including the datasets, error metrics, benchmarks and statistical testing. An analysis of the results is presented in Section \ref{sec:results}. Section \ref{sec:concluson} concludes the paper and discusses possible future research.

\section{Related work}
\label{sec:related_work}

In the following, we discuss the related prior work in the areas of TAR models, GFMs and the state-of-the-art tree-based forecasting models.

\subsection{Threshold autoregressive models}
\label{sec: tar_models}
TAR models \citep{tong_1977_threshold,tong_1993_setar} are piece-wise linear models that model the state space of a given prediction problem using multiple AR models. TAR models are also known as AR regime switching models which use separate (same or different order) AR functions to model different regimes. 

TAR models have variants such as SETAR and STAR \citep{Terasvirta_1994_star} which are different in terms of the transition function and the method of defining regimes. The simplest version of the TAR model is known as the SETAR model \citep{tong_1993_setar} which defines two or more regimes based on a particular lagged value of the time series itself. Thus, SETAR models are required to estimate the optimal lag, and the optimal threshold corresponding with that chosen optimal lag to define regimes. The most common way of defining these optimal lag and threshold values is using a grid search. The optimal thresholds can be also identified by analysing the places of time series showing significant increasing or decreasing rates \citep{ghosh_2006_self} and by using memetic algorithms \citep{bergmeir_2012_memetic}. The transition function used by the SETAR model is generally a step function. Thus, during prediction, SETAR models use only one AR model. Versions of the SETAR model exist that use statistical tests such as a Lagrange Multiplier (LM) test \citep{breusch_1980_lagrange} to determine the linearity captured by the AR models \citep{Terasvirta_1994_star}. In STAR models \citep{Terasvirta_1994_star}, the transition can happen either using a past value of a series or an external variable. Unlike the SETAR models, STAR models use a transition function such as exponential or logistic, resulting in ESTAR and LSTAR models, respectively, and thus, the transition happens smoothly. During prediction, STAR models use multiple AR models to determine the final output where the corresponding weight of each contributing AR model is determined based on the transition function.

\citet{Aznarte_5556012} and \citet{aznarte_00506473} establish the equivalence between TAR models and certain types of fuzzy rule-based systems. Inspired by this work, we exploit equivalences between TAR models and decision trees, and extend TAR models to global hierarchical models which train across series.

TAR models have been used to address many real-world forecasting problems such as stock market forecasting \citep{NARAYAN2006103}, exchange rate forecasting \citep{Pippenger_1998_exchange} and electricity price forecasting \citep{rambharat_2005_threshold}. To the best of our knowledge, none of these TAR models have been used as global models across time series yet.

\subsection{Global time series forecasting}
\label{sec: gfms}
Global time series forecasting \citep{ref_105} is a relatively recent trend in forecasting that became popular after contributing to the winning methods of the M4 \citep{ref_30} and M5 \citep{makridakis_2020_m5} forecasting competitions. In particular, many top solutions of the M5 competition used global (tree-based) models. GFMs build a single forecasting model across many series sharing the same parameters across all series. Thus, the model has the capability to explore cross-series information with a fewer amount of parameters compared to traditional univariate forecasting models. 

Many works use Recurrent Neural Networks \citep[RNN,][]{ref_6} as the baseline model when developing global forecasting frameworks. The winning method of the M4 competition \citep{ref_1} uses an ensembling approach, ensemble of specialists \citep{ref_1}, that combines the predictions of a set of globally trained RNNs. \citet{ref_23} use globally trained RNNs for forecasting. One of the state-of-the-art probabilistic forecasting algorithms, DeepAR \citep{ref_99}, uses a global AR RNN. \citet{godahewa_2020_weekly} propose a weekly forecasting framework that uses global RNNs. 

In particular, the winning method of the M5 forecasting competition uses a globally trained LightGBM model \citep{guolin_lightgbm_2017}. GFMs have also contributed to the winning methods of many other recently held Kaggle competitions \citep{BOJER2020}. Nowadays, GFMs are used with many real-world applications such as energy optimisation \citep{ref_12}, sales demand forecasting \citep{ref_13} and emergency medical services demand forecasting \citep{Bandara2020-en}. 

\citet{godahewa_2021_ensembling} and \citet{ref_2} show that a GFM should be trained with a set of related time series. Those authors further claim that if a collection of time series has heterogeneous series, then the series should be first grouped based on their similarity and then, separate GFMs should be trained per each series group. In line with this concept, our proposed SETAR-Tree model trains multiple GFMs, one per leaf node, because the instances belonging to a particular leaf node are similar in terms of their values corresponding with the optimal lag identified using the SETAR methodology. As the baseline GFM, we use a linear AR model trained across a set of time series that is known as a PR model \citep{gelman_hill_2006}.

\subsection{Tree-based forecasting models}
\label{sec: trees}
There are many popular tree-based algorithms used for forecasting, including regression trees \citep{loh_2011_classification}, gradient boosted trees \citep{friedman_2001_greedy}, XGBoost \citep{chen_xgboost_2016}, CatBoost \citep{NEURIPS2018_14491b75}, and LightGBM \citep{guolin_lightgbm_2017}.

As forecasting is usually a regression task with numerical values, regression trees \citep{loh_2011_classification} are used for forecasting, as opposed to decision trees for classification. \citet{spiliotis_2022_trees} provides an overview of how decision trees can be used for time series forecasting. In particular, the author highlights that in the forecasting domain, a regression tree provides a piece-wise approximation to a single continuous function in standard regression. Regression trees use every possible binary split on every input attribute to determine the node splits. Regression trees have been successfully applied to address real-world time series forecasting applications. \citet{ref_53} use bagged regression trees to handle the diverse dynamic regimes and non-stationarities observed in real-world time series. \citet{cerqueira2019arbitrage} use a regression tree as a base model of a pool of forecasting experts where the forecasts of experts are dynamically combined using arbitrating to obtain the final forecasts. Gradient boosted trees \citep{friedman_2001_greedy} use a collection of regression trees as a sequential ensemble model. The first regression tree is trained on the actual observations of the training instances where the remaining trees are trained to predict the residuals of the previous tree in the sequence. XGBoost \citep{chen_xgboost_2016} is an accurate, efficient and scalable version of gradient boosted trees. To define the best node splits, XGBoost uses similarity scores in a way that the gain is maximised. Furthermore, to reduce overfitting, XGBoost uses regularisation parameters when calculating similarity scores and tree pruning. The second winning method of the M4 forecasting competition, Feature-based Forecast Model Averaging \citep[FFORMA,][]{ref_3} uses XGBoost as a meta-learner to determine the weights that should be used to combine the predictions of a set of base forecasting models. 
CatBoost \citep{NEURIPS2018_14491b75} considers the order of data points during modelling and thus, it is more suitable to address time series forecasting problems. CatBoost more effectively deals with categorical variables compared to other tree-based algorithms.

As stated earlier, tree-based forecasting models recently obtained popularity in the forecasting field after contributing to the winning approach and many other top submissions in the M5 forecasting competition \citep{makridakis_2020_m5}. In particular, many top submissions of the competition incorporate a highly efficient gradient boosted tree-based model, LightGBM \citep{guolin_lightgbm_2017}. The LightGBM trees use the leaf-wise tree growth approach instead of the level-wise tree growth used in traditional tree-based algorithms. It identifies the best node splits by using Gradient-based One-Side Sampling (GOSS). LightGBM can handle extremely large datasets as well and nowadays, it is used with many real-world forecasting applications such as cryptocurrency price trend forecasting \citep{SUN2020101084}, sales demand forecasting \citep{weng_2020_supply} and wind power forecasting \citep{ju_2019_model}. \citet{JANUSCHOWSKI2021} highlight that tree-based algorithms are blackbox learners which are highly efficient and more robust than the existing deep learning techniques. In particular, the LightGBM algorithm offers a large number of loss functions and missing value handling methods where the users can choose the corresponding options based on their application.

Even though these tree algorithms are efficient, they are not tailored to the forecasting problem. For example, they consider the average of the training outputs in a particular leaf node as the forecast of the test instances that reach that node during forecasting. In contrast, our proposed SETAR-Tree algorithm trains a global linear AR model in each leaf node, allowing each leaf node to learn the cross-series information. 

In the literature, there already exist tree models which train linear models in leaf nodes, known as \textit{linear regression model trees} \citep{quinlan1992learning}. They analyse every possible split to determine the node splits where the split which maximises the difference between the standard deviation of the actual target values of the parent node and the mean of the standard deviations of the actual target values of child nodes, is considered as the optimal split. The prediction for a given test instance is obtained by its corresponding leaf node where the prediction is then adjusted based on the nodes from the root node to its corresponding leaf node. Cubist \citep{kuhn_2013_applied} is such a linear model tree that performs rule-based modelling. It trains linear models in all tree nodes where the predictions are obtained from the leaf nodes and are smoothed afterwards. \citet{Lefakis_2019_efficient} propose a piecewise linear regression tree which trains regularised linear models in leaf nodes. The nodes are split in a way that the least square errors of linear predictors after splitting are minimised. \citet{loh_2002_regression} proposes a regression tree which uses the significance of the chi-square test to identify the variables that should be used during node splitting. \citet{dutang_2022_explicit} propose a splitting procedure for regression trees where the optimal splits are chosen based on the explicit likelihood. \citet{zeileis_2008_model} propose a regression model which uses recursive partitioning where the nodes are only further split if there exists an overall parameter instability. \citet{hothorn_2006_ctree} propose CTree, a regression model that uses recursive partitioning with conditional inference procedures. There are hardly any works in the literature that propose tree-based models particularly for forecasting that we are aware of, and the few works available address local forecasting, not global forecasting models. In particular, the STR-Tree proposed by \citet{rosa_str_2008}, though proposed for regression, is deemed by its authors to be trivially extendable to forecasting, and \citet{epprecht_2012_starx} propose a tree method to predict stock market returns where in the leaf nodes AR models with extra inputs are trained. These works use the concept of STAR models during node splitting, which is similar to our approach but limits them effectively to smaller sample sizes, as the smooth transitions make their model training computationally considerably more costly. Thus, these methods are inherently more suitable for local forecasting and using them in current typical global forecasting scenarios brings scalability issues.

Our proposed SETAR-Tree algorithm is in particular tailored to the time series forecasting task by using the concept of SETAR models in defining splits where the optimal lag and the threshold that should be used for node splitting is determined using a grid search approach, without considering every possible split as used by the above explained linear regression model trees. Our grid search approach selects the lag and the threshold that is corresponding with the split which minimises the Sum of Squared Errors (SSE) at the child nodes. Furthermore, to determine the validity of a split, we use a statistical linearity test and the error reduction percentage that can be gained from the split, where a node is only further split if making that split is worth enough. In leaf nodes, we train global AR models allowing the models to learn the cross-series information and they do not use any regularisation during model fitting. Also, the predictions for the test instances are obtained only from their corresponding leaf nodes where no further adjustments are applied to the predictions as in linear regression model trees (for details, see Section \ref{sec: setar_tree_model}). 

A collection of parallel executed tree models is known as a forest. The RF \citep{breiman_2001_random} algorithm is the most common type of forest algorithms which averages the predictions provided by a set of diverse regression trees to obtain the final predictions. The trees can be diverse in terms of the data and features used during training. This ensembling technique is also known as bagging in the machine learning literature and it is expected to improve the accuracy of the final predictions compared to the predictions of the individual trees by properly addressing data, model and parameter uncertainties \citep{PETROPOULOS2018545, BERGMEIR2016303}. Also, \citet{philippe_2021_bag} argues that the RF does an optimal pruning as the splits that overfit are averaged out through bagging. In the context of linear model trees, the Macroeconomic RF (MRF) proposed by \citet{philippe_2020_rf} consists of a collection of linear model trees that use the concept of TAR models during node splitting. However, MRF is not designed for global time series forecasting. Unlike our SETAR-Tree, the individual trees in MRF use regularisation during parameter estimation, and use moving average factors calculated using the lagged values of time series as regressors during training. The trees also do not use any stopping criterion based on statistical tests or error reduction percentages. In particular, the individual trees in MRF do not use early stopping where they are allowed to grow into higher depths ensuring the trees are sufficiently diversified.  Furthermore, \citet{athey_2019_generalised} propose generalized RFs, a method based on RFs that can be used for non-parametric statistical estimation. Consequently, we propose a forest model that uses a collection of SETAR-Trees where the individual trees are diverse in terms of the significance of the linearity test and the error threshold used to make the node splits (for details, see Sections \ref{sec: lin_test}, \ref{sec: error_imp} and \ref{sec: setar_forest}). 

\section{Methodology}
\label{sec:methodology}
This section first gives a brief overview of the theory of SETAR models. Then, we explain the terminology of GFMs. Later, the proposed SETAR-Tree and forest algorithms are explained in detail.

\subsection{Self exciting threshold autoregressive models}
\label{sec: setar}
For our study, we consider a 2-regime SETAR model \citep{tong_1993_setar} which is defined in Equation \ref{eqn: setar_def}. Here, $y_t$ is the value of the series at time $t$, $n$ is the number of past time series lags, $T$ is the threshold value, and $\beta^{i}$ and $\epsilon^{i}$ are respectively the AR parameters and error terms of the $i^{th}$ regime.

\begin{equation}
\label{eqn: setar_def}
    y_t =
    \begin{cases}
      \sum_{l=1}^{n} \beta^{(1)}_{l}{y_{t-l}} + \epsilon^{(1)}_t,& \text{if } y_{t-l} < T \\
      \\
      \sum_{l=1}^{n} \beta^{(2)}_{l}{y_{t-l}} + \epsilon^{(2)}_t,& \text{otherwise} 
    \end{cases}
\end{equation}

As shown in Equation \ref{eqn: setar_def}, SETAR models define regimes based on a particular lagged value of the time series itself. One regime is defined for the training instances where the corresponding lagged value is less than the threshold value $T$, and the other regime is defined for the remaining instances where the corresponding lagged value is greater than or equal to $T$. Thus, SETAR models have as hyperparameters the optimal lag, $l$, and the optimal threshold value, $T$ corresponding with $l$ to define the regimes.

\subsection{Terminology of global forecasting models}
\label{sec: terminology}
For our study, we consider univariate time series forecasting which predicts the future values of a particular time series using its own past values and possibly, some external covariates. We use the usual out-of-sample evaluation in forecasting that reserves a block of data from the very end of the time series for evaluation \citep{BERGMEIR2012192}. For that, the time series are first split into training and test parts where the training parts are used for model training and the test parts are the actual values corresponding with the expected forecast horizon. 

\begin{figure}
\centering
  \includegraphics[scale=0.6]{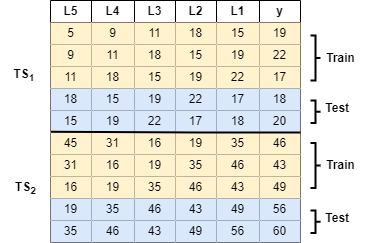}
  \caption{Visualisation of an embedded matrix created using two time series: $TS_{1}$ and $TS_{2}$. Each row contains a window of a time series which is composed with a set of lagged values ($L1$ to $L5$) and the corresponding next value/true output ($y$). The windows corresponding with the training and test sets are respectively highlighted in yellow and blue.}
  \label{fig: time_series}
\end{figure}

Let $TS_{1}$ and $TS_{2}$ be two time series. To train purely autoregressive GFMs, as shown in Figure \ref{fig: time_series}, an embedded matrix is constructed using $TS_{1}$ and $TS_{2}$ where each row contains a window of a time series which is composed of a set of lagged values and the corresponding next value. The rows of the embedded matrix that are corresponding with the training parts of the series are used to train GFMs. The forecasts corresponding with the expected forecast horizon are obtained for each series using the trained GFMs. During the evaluation, the forecasts are compared with their corresponding actual values that are in the test set and the forecasting errors are measured using the error metrics explained in Section \ref{sec:error_metrics}. 

\subsection{SETAR-Tree model}
\label{sec: setar_tree_model}
\begin{figure}
\centering
  \includegraphics[width=\textwidth]{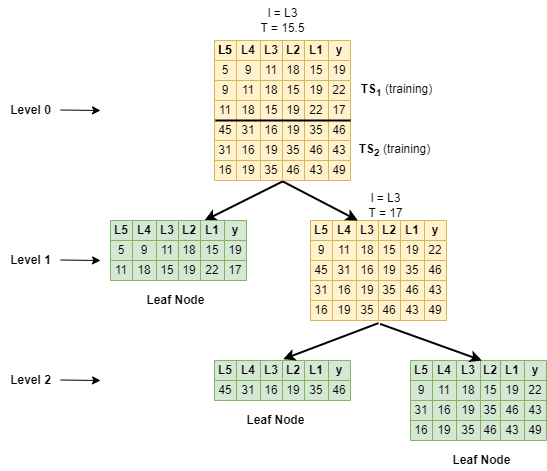}
  \caption{Overview of an example SETAR-Tree constructed using $TS_{1}$ and $TS_{2}$. The matrices containing the past time series lags and their corresponding true outputs are considered as nodes. Each row of a matrix is considered as a training instance. The training instances of each node are optimally split into child nodes based on the concept of SETAR models. Here $l$ and $T$ mentioned with the split nodes refer to the optimal lag and the optimal threshold that are used to split at a particular node. A node is only split if there exists remaining non-linearity in the training instances of that node and/or it can gain a certain training error reduction by splitting. This example tree model has three levels. This model uses a leaf-wise tree growth approach and thus, the branches of the tree may have different lengths}
  \label{fig: setar_tree}
\end{figure}

Figure \ref{fig: setar_tree} shows an overview of an example SETAR-Tree model. Level 0 refers to the root node of the tree. The root node is an embedded matrix constructed using the training parts of $TS_{1}$ and $TS_{2}$ as explained in Section \ref{sec: terminology}. Each row of the embedded matrix is used as a training instance. We use the concept of SETAR models to split the training instances into child nodes. As we only consider 2-regime SETAR models, a given tree node is split into two child nodes. The optimal lag, $l$ and the optimal threshold, $T$ that are used to split each node are determined by using a grid search approach as explained in Section \ref{sec: opt_lag_th}. The instances where the corresponding value of the Lag $l$ is less than $T$, and greater than or equal to $T$, are separately grouped into child nodes. However, a node is only split if making that split is worth enough. In particular, a node is only split if there exists remaining non-linearity in the training instances of that node (Section \ref{sec: lin_test}) and/or it can gain a certain training error reduction by splitting (Section \ref{sec: error_imp}). Thus, our model uses the leaf-wise tree growth approach where the branches of the tree may have different lengths. For an example, at Level 1, only one node is further split into child nodes. The other node is not split and it has been treated as a leaf node because further splitting that node does not fulfill the specified criteria.

The two methods we use to determine whether a node should be split or not, a statistical linearity test and the error reduction gained by node splitting, are explained in detail in Sections \ref{sec: lin_test} and \ref{sec: error_imp}. They control the growth of the tree and thus, the model can itself determine the maximum tree depth that it should allow. In particular, these two methods act as the stopping criteria of the tree growth where constructing the tree is completed when none of the nodes should be further split. The leaf nodes can be there at different tree levels as shown in Figure \ref{fig: setar_tree} where one leaf node is in Level 1 and two leaf nodes are in Level 2. In each leaf node, a global PR model is trained using the instances corresponding with that node.

During testing, the model identifies the leaf node corresponding with a given test instance by following the same optimal lags and thresholds it previously used during node splitting starting from the root node. The prediction of the test instance is then obtained using the trained PR model corresponding with its leaf node.

The SETAR-Tree requires minimal hyperparameter tuning. A few parameters are used for the statistical linearity test and for measuring error reduction in node spitting. However, the model provides competitive results with their default parameters that we provide and thus, setting externally tuned hyperparameters is not mandatory for our model.

\subsubsection{Finding optimal lags and thresholds}
\label{sec: opt_lag_th}
To find the lag and threshold that will be used to split a given node, we utilise a standard grid-search approach. Given a node, we consider each lag and a set (grid) of thresholds with which to split the node into two children. For each lag and threshold pair in the grid, separate linear models are fitted to the two subsets of data formed by  partitioning the chosen lag at the specified threshold. The lag and the threshold that are results in the split with the minimum SSE at the child nodes is selected as the ``optimal'' lag and threshold for splitting. 

To increase the efficiency of the grid-search, we exploit the properties of the least-squares estimates during fitting. Recall that the leaf models are linear AR models of the form
\[
    y = X \beta + \varepsilon
\]
where $X$ is the lag-matrix, $y$ are the outputs, $\beta$ the AR coefficients and $\varepsilon$ a vector of errors. The least-squares estimates of $\beta$ are well known to be
\begin{equation}
    \label{eq:LS:estimates}
    \hat{\beta} = (X^{T} X)^{-1} X^{T} y
\end{equation}
where T and -1 respectively refer to the transpose and inverse of the matrix. The SSE of the least-squares fit is given by
\begin{equation}
    \label{eq:RSS:LS}
    {\rm SSE}(\hat{\beta}) = y^T y - \hat{\beta}^T X^T X \hat{\beta}.
\end{equation}
An important property of the least-squares estimates are that they depend on $X$ and $y$ only through their inner products. We can exploit this to fit the  models for all potential thresholds in $O(n \log n)$ time. Let $X_l$ be the lag variable we will be splitting on, and let $s = (s_1,\ldots,s_q)$ be the $q$ candidate thresholds for splitting. Define the sets
\[
    S_i = \left\{ j : x_{j,l} \leq s_i \right\}, (i = 1,\ldots,q) \; {\rm and} \; S_{q+1} = \left\{ j : x_{j,l} > s_q \right\}.
\]
These sets indicate which rows fall into each of the $q+1$ different partitions of the data defined by the set of thresholds. Let $\bar{x}_i$ denote the $i$-th row of $X$; we can then define the quantities
\begin{equation}
    \label{eq:inner:products}
    B_{(k)} = \sum_{i=1}^k \sum_{j \in S_i} \bar{x}_{j} \bar{x}_{j}^{T}, \; \; c_{(k)} = \sum_{i=1}^k \sum_{j \in S_i} \bar{x}_{j} y_j \; \; {\rm and}\; \; d_{(k)} = \sum_{i=1}^k \sum_{j \in S_i} y_j^2.
\end{equation}
These are the relevant inner products for the ``left'' child node model formed by splitting the data at each of the thresholds $s_k$. From Equation \ref{eq:LS:estimates}, if we choose to split at threshold $s_k$, the least squares estimates for the left and right child node models can be written as
\begin{equation}
    \label{eq:LS:est}
    \hat{\beta}{(L_k)} = B_{(k)}^{-1} c_{(k)} \; \; {\rm and} \; \;
    \hat{\beta}{(R_k)} = \left( X^{T} X - B_{(k)} \right)^{-1} \left( X^{T} y - c_{(k)} \right)
\end{equation}
respectively. Using Equation \ref{eq:RSS:LS}, the SSE in the left and right child node models can be then written as
\begin{equation}
    \label{eq:RSS:L}
    {\rm SSE}{(L_k)} = d_{(k)} - \left( \hat{\beta}{(L_k)} \right)^{T} B_{(k)} \left( \hat{\beta}{(L_k)} \right)
\end{equation}
and
\begin{equation}
    \label{eq:RSS:R}
    {\rm SSE}{(R_k)} = \left( y^{T} y - d_{(k)} \right) - \left( \hat{\beta}{(L_k)} \right)^{T} \left( X^{T} X - B_{(k)} \right) \left( \hat{\beta}{(L_k)} \right)
\end{equation}
respectively. When considering a lag $x_l$ to split on, we can first sort the rows of $X$ by $x_l$. In practice, these sorted values can then be used to find a set of suitable thresholds; typically we choose fifteen equispaced quantiles of the vector $x_l$. We can then use Equation \ref{eq:inner:products} to compute the relevant inner products for the first threshold, i.e., $B_{(1)}$, $c_{(1)}$ and $d_{(1)}$. We can then iteratively try each potential threshold $s_1,\ldots,s_q$ in turn using Equations \ref{eq:LS:est}, \ref{eq:RSS:L} and \ref{eq:RSS:R}, exploiting the fact that we can update the inner products using
\[
    B_{(k+1)} = B_{(k)} + \sum_{j \in S_k} \bar{x}_j \bar{x}_j^T,
\]
as we move from threshold $s_k$ to $s_{k+1}$, with similar expressions for $c_{(k)}$ and $d_{(k)}$ to efficiently compute the running totals of these quantities. 

\subsubsection{Statistical linearity test}
\label{sec: lin_test}
The usage of linearity tests in determining the amount of regimes and the goodness of TAR models is well-established in the econometrics and statistics literature \citep{Terasvirta_1994_star}. However, to the best of our knowledge, the linearity tests have not been used to determine the goodness of a SETAR model trained across series, in particular to determine the depth of a hierarchical SETAR model as we propose.

We use the general linear F-test \citep{Box_1953_tests} to determine whether there exists a remaining non-linearity of a set of training instances in a particular tree node. The node is only further split if there exists a significant remaining non-linearity. 

The null hypothesis of the linear F-test, $H_0$ states that there exists no significant remaining non-linearity in the training instances and thus, the parent node is enough to model the instances. The alternative hypothesis, $H_1$ states that there exists a significant remaining non-linearity in the training instances and therefore, a node should be further split into child nodes.

In particular, the linear F-test calculates a statistic, namely the F-statistic, which compares a complex model and a reduced model, and determines whether the reduced model should be rejected in favour of the complex model. In our node splitting scenario, the reduced and complex models respectively refer to the global PR models that are trained on parent and child nodes corresponding with a particular split. The suitability of the split is determined based on the value of the F-statistic calculated using the SSE corresponding with PR models built on parent and child nodes.

The SSE of a parent node and the corresponding child nodes are respectively defined in Equations \ref{eqn:parent_sse} and \ref{eqn:child_sse}. Here, $N$ is the number of training instances, $y_k$ are the observed values, $yp_k$ are the fitted values of the parent node model and $yc_k$ are the fitted values of the child node models. 

\begin{equation}
\label{eqn:parent_sse}
    SSE(P) = \sum_{k=1}^{N} ({y_{k} - yp_{k}})^2
\end{equation}

\begin{equation}
\label{eqn:child_sse}
    SSE(C) = \sum_{k=1}^{N} ({y_{k} - yc_{k}})^2
\end{equation}

The F-statistic corresponding with a node split is then calculated using Equation \ref{eqn:f_stat} where $df_P$ and $df_C$ are the degrees of freedom of parent and child node models, respectively.

\begin{equation}
\label{eqn:f_stat}
    F^* = \left(\frac{SSE(P) - SSE(C)}{df_P - df_C}\right)  {\left(\frac{SSE(C)}{df_C}\right)}^{-1}
\end{equation}

For simplicity, we consider the same number of past lag values when training the PR models in all nodes. Hence, $df_P$ and $df_C$ are defined as in Equations \ref{eqn:dfp} and \ref{eqn:dfc} where $L$ is the number of past lags used for modelling. 

\begin{equation}
\label{eqn:dfp}
    df_P = N - L - 1
\end{equation}

\begin{equation}
\label{eqn:dfc}
    df_C = N - 2L - 2
\end{equation}

Substituting $df_P$ and $df_C$ reduces Equation \ref{eqn:f_stat} as follows.

\begin{equation}
\label{eqn:f_stat_reduced}
    F^* = \left(\frac{SSE(P) - SSE(C)}{L + 1}\right) {\left(\frac{SSE(C)}{N - 2L - 2}\right)}^{-1}
\end{equation}

The $p$-value of the F-test is determined by comparing the value of $F^*$ to the $F$ distribution. If the $p$-value is less than the considered significance level, $\alpha$, then $H_1$ is accepted and the node is further split into child nodes. Otherwise, the node is kept as a leaf node of the tree.

As a form of multiple testing correction \citep{noble_2009_multiple}, we gradually decrease the value of $\alpha$ when the tree grows level by level as defined in Equation \ref{eqn:alpha}, where $\alpha_{d}$ and $\alpha_{d+1}$ are the significance levels corresponding with the tree levels $d$ and $d+1$, respectively.

\begin{equation}
\label{eqn:alpha}
    \alpha_{d+1} = \frac{\alpha_{d}}{\mathit{significance\_divider}}
\end{equation}

This reduces unnecessary tree growth and acts as a way of preventing model overfitting. The values of $\alpha_0$ and $\mathit{significance\_divider}$ are user defined, with default values of 0.05 and 2, repectively. As our trees perform binary splits, using a $\mathit{significance\_divider}$ of 2 corresponds to a Bonferroni correction, where the family-wise error is kept constant across the levels of the tree. Larger values lead accordingly to more conservative significance levels.

\subsubsection{Error reduction in node splitting}
\label{sec: error_imp}
We also consider the error reduction percentage to determine whether a node should be further split or not. The child nodes models together use more parameters during model fitting and thus, they always give lower training error compared to the parent node model. Here, the purpose is to check whether the error reduction percentage that can be gained by splitting a node is considerably high or not. A node is only split if the error reduction percentage is greater than or equal to a particular error threshold ($e_t$) and otherwise, the node is considered as a leaf node of the tree. The value of $e_t$ is user defined and based on our preliminary experiments, it is set to 3\%, by default.

\bigskip

Algorithm \ref{alg1} shows the training and testing phases of the SETAR-Tree model. The algorithm takes the following parameters as inputs.

\begin{algorithm}[!htb]
\fontsize{7}{7}
\caption{SETAR-Tree Algorithm}
\label{alg1}
\begin{algorithmic}[1]

\Procedure{setar\_tree}{$training\_set$, $forecast\_horizon$, $lag$, $stopping\_criteria$, $max\_depth=1000$ $alpha0=0.05$, $significance\_divider=2$, $error\_threshold=0.03$} 

\State $alpha$ $\gets$ $alpha0$
\State $tree$ $\gets$ $list()$
\State $opt\_lags$ $\gets$ $list()$
\State $opt\_thresholds$ $\gets$ $list()$
\State $input\_matrix$ $\gets$ $create\_input\_matrix(training\_set, lag)$ 
\State $node\_data$ $\gets$ $input\_matrix$

\For{$depth$ in $max\_depth$}

\State $level\_significant\_node\_count$ $\gets$ 0

  \For{$node$ in $node\_data$}
     \State $opt\_params$ $\gets$ $get\_opt\_params(node)$
     \State $opt\_lag$ $\gets$ $opt\_params[1]$
     \State $opt\_threshold$ $\gets$ $opt\_params[2]$
     \State $child\_nodes$ $\gets$ $split\_node(node, opt\_lag, opt\_threshold)$
     
     \If{$stopping\_criteria$ $=$ $"linearity\_test"$} 
		\State $good\_split$ $\gets$ $check\_linearity(node, child\_nodes, alpha)$ 
	 \ElsIf{$stopping\_criteria$ $=$ $"error\_reduction"$} 
	    \State $good\_split$ $\gets$ $check\_error\_reduction(node, child\_nodes, error\_threshold)$ 
	 \Else 
	    \State $good\_split$ $\gets$ $check\_linearity(node, child\_nodes, alpha)$ \textbf{and} \\ $check\_error\_reduction(node, child\_nodes, error\_threshold)$ 
	 \EndIf

	 \If{$good\_split$}
	    \State $tree[depth]$ $\gets$ $add(tree[depth], child\_nodes)$
	    \State $opt\_lags[depth]$ $\gets$ $add(opt\_lags[depth], opt\_lag)$
	    \State $opt\_thresholds[depth]$ $\gets$ $add(opt\_thresholds[depth], opt\_threshold)$
	    \State $level\_significant\_node\_count$ $\gets$ $level\_significant\_node\_count + 1 $
	 \EndIf
  \EndFor
  
  \If{$level\_significant\_node\_count > 0$}
	    \State $node\_data$ $\gets$ $tree[depth]$
	    \State $alpha$ $\gets$ $alpha/significance\_divider$
  \Else
      \State $break$
  \EndIf
\EndFor

\State $leaf\_nodes$ $\gets$ $get\_leaf\_nodes(tree)$
\State $leaf\_models$ $\gets$ $list()$

\For{$leaf$ in $leaf\_nodes$}
   \State $model$ $\gets$ $train\_pr\_model(leaf)$  
   \State $leaf\_models$ $\gets$ $add(leaf\_models, model)$
\EndFor

\algstore{myalg}
\end{algorithmic}
\end{algorithm}

\begin{algorithm}  
\fontsize{7}{7}                   
\begin{algorithmic} [1]                   
\algrestore{myalg}

\State $predictions$ $\gets$ $matrix()$
\State $test\_set$ $\gets$ $create\_test\_set(training\_set, lag)$ 

\For{$f$ in $forecast\_horizon$}
\For{$s$ in $nrow(test\_set)$}
   \State $leaf\_node\_index$ $\gets$ $find\_leaf\_node\_index(test\_set[s,], tree, opt\_lags, opt\_thresholds)$  
   \State $predictions[s,f]$ $\gets$ $predict(leaf\_models[leaf\_node\_index], test\_set[s,])$
\EndFor
\State $test\_set$ $\gets$ $update\_test\_set(test\_set, predictions[,f])$ 
\EndFor

\Return $predictions$
\EndProcedure
\end{algorithmic}
\end{algorithm}

\begin{itemize} 
    \item $\textbf{training\_set}$: contains the training parts of each series of the dataset
    \item $\textbf{forecast\_horizon}$: the length of the prediction period
    \item $\textbf{lag}$: the number of past lags used to forecast the next series value
    \item $\textbf{stopping\_criteria}$: the criteria that are used to determine whether a node should be split
    \item $\textbf{max\_depth}$: maximum possible tree depth (optional)
    \item $\textbf{alpha0}$: the initial significance level to be used by the F-test (optional)
    \item $\textbf{significance\_divider}$: the significance level used by the F-test is sequentially divided at each tree level by this value (optional)
    \item $\textbf{error\_threshold}$: minimum error reduction that should be gained to split a node (optional)
\end{itemize}

The algorithm first creates an embedded matrix using the training parts of each series which is considered as the root node of the tree. The function \textit{create\_input\_matrix} creates this embedded matrix taking \textit{training\_set} and \textit{lag} as inputs (line 6). It then finds the optimal lag and the threshold that should be used to split the node into child nodes using the function \textit{get\_opt\_params} (line 11). This function internally conducts a grid search over the lags of the node and selects the lag and the threshold that are corresponding with the split which provides the minimum SSE at the child nodes. Then, the function \textit{split\_node} (line 14) splits the parent node into child nodes using the optimal lag (\textit{opt\_lag}) and the threshold (\textit{opt\_threshold}) returned by \textit{get\_opt\_params}. Based on the \textit{stopping\_criteria}, the split is further assessed to determine whether the child nodes should be added to the tree or not. The functions \textit{check\_linearity} (line 16) and \textit{check\_error\_reduction} (line 18) respectively check whether there exists remaining non-linearity in the training instances of the parent node and whether the training error reduction gain by node splitting is greater than the \textit{error\_threshold}. If the split is worth to make, then the child nodes are added to the tree (line 24). The tree building procedure is stopped if none of the nodes at the current tree level is split (line 34). If at least one node at the current tree level is split, then the same node splitting procedure is conducted for the next tree level. At each tree level, the significance level used by the F-test, \textit{alpha}, is decreased by dividing it by the \textit{significance\_divider} (line 32). After completing the tree construction process, our method trains a global PR model at each leaf node using the function \textit{train\_pr\_model} (line 40). During the testing phase, it first creates a set of test instances, \textit{test\_set}, using the function \textit{create\_test\_set} which returns the last set of lags in each training series (line 44). The forecasts for the required forecast horizon are then obtained iteratively. In each iteration, the function \textit{find\_leaf\_node\_index} (line 47) identifies the corresponding leaf node for each test instance by following the same optimal lags (\textit{opt\_lags}) and thresholds (\textit{opt\_thresholds}) it previously used during node splitting. The predictions for the test instances are then obtained using the corresponding trained models at their leaf nodes (line 48). In each iteration, the \textit{test\_set} is also updated using the function \textit{update\_test\_set}, by appending the forecasts of the previous iteration (line 50).

\subsubsection{Training with covariates}
\label{sec: covariates}
The proposed SETAR-Tree model can also be trained with external numerical and categorical covariates in addition to the past lags of time series. The numerical covariates are treated in the same way as the time series lagged values. The categorical covariates are converted into numerical format beforehand by applying one-hot encoding which then allows to treat them in the same way as numerical attributes. When determining the optimal attribute and the threshold to split a node, the past lags of series, numerical covariates and categorical covariates are all considered together.

\subsection{SETAR-Forest model}
\label{sec: setar_forest}
A forest is an ensemble model that aggregates the predictions provided by a set of diverse tree models to obtain a prediction, often through bootstrap aggregation (``bagging'') \citep{breiman_2001_random}.

We introduce a new forest algorithm, \textit{SETAR-Forest}, which consists of a collection of diverse SETAR-Trees. The trees are made diverse by varying the initial significance level ($\alpha_0$), the  significance divider used to calculate the sequence significance levels and the error reduction percentage threshold ($e_t$) used during node splitting. In a SETAR-Tree, these three parameters are set by default based on our preliminary experiments and in contrast to that, in the forest algorithm, each tree selects the values of these parameters randomly. For our experiments, we consider ten SETAR-Trees for a SETAR-Forest. An individual SETAR-Tree does not use bagging whereas the SETAR-Forest does. A set of randomly chosen 80\% of instances and all attributes of them are used to train each SETAR-Tree in a SETAR-Forest. Thus, the values of bagging frequency, bagging fraction and feature fraction of our SETAR-Forest are 10, 0.8, and 1, respectively, by default. However, our implementation takes these values as external parameters and thus, users can modify them based on the application.

During testing, the forecasts of the test instances are obtained by averaging the corresponding forecasts provided by all SETAR-Trees in the forest.

\section{Experimental framework}
\label{sec:experiments}
In this section, we discuss the experimental datasets, error metrics, and benchmarks used in our experiments.

\subsection{Datasets}
\label{sec:datasets}

\begin{table*}
\centering
\caption{Datasets information}
\resizebox{1.0\textwidth}{!}{
\begin{tabular}{lccccc}
\hline
\textbf{Dataset Name} & \textbf{No. of} & \textbf{Forecast} & \textbf{Frequency} & \textbf{Minimum} & \textbf{Maximum} \\
 & \textbf{Time Series} & \textbf{Horizon} &  & \textbf{Length} & \textbf{Length} \\
\hline
Rossmann &  1115 & 48 & Daily & 894 & 894 \\
Wikipedia Web Traffic & 1000 & 59 & Daily & 744 & 744\\
Favorita &  1000 & 16 & Daily & 1668 & 1668\\
M5 &  1490 & 28 & Daily & 1941 & 1941\\
Tourism Monthly & 366 & 24 & Monthly & 67 & 309 \\
Tourism Quarterly & 427 & 8 & Quarterly & 22 & 122 \\
Chaotic Logistic & 100 & 8 & - & 592 & 592 \\
Mackey-Glass & 100 & 8 & - & 592 & 592 \\
\hline
\end{tabular}}
\label{tab:dataset_overview}
\end{table*}

We use eight experimental datasets\footnote{The experimental datasets are available at \url{https://github.com/rakshitha123/SETAR_Trees/tree/master/datasets}.}, six publicly available datasets and two simulated datasets, to evaluate the performance of our proposed SETAR-Tree and forest algorithms. The datasets are briefly explained in the following and Table \ref{tab:dataset_overview} provides their summary statistics.

\begin{itemize}    
    \item Rossmann Sales Dataset \citep{rossmann_2015_kaggle}: The dataset from the Rossmann Sales forecasting competition that shows the daily sales of a set of Rossmann stores. The missing observations of this dataset are replaced by carrying forward the corresponding last observations (LOCF method). Several covariates are also available with this dataset such as the daily number of customers to each store, status of the stores (open or closed), promotion details and state/school holidays information. 
    
    \item Kaggle Wikipedia Web Traffic Dataset \citep{ref_32}: The first 1000 time series from the Kaggle Wikipedia Web Traffic forecasting competition that shows the number of daily hits for a given set of Wikipedia web pages. Following the procedure suggested by \citet{ref_6}, the missing observations of this dataset are replaced by zeros. Covariates are also available with this dataset such as day of the week. The dataset was extracted from \citet{godahewa_2021_monash}.
    
    \item Favorita Dataset \citep{favourita_2018_kaggle}:  The first 1000 time series from the Corporaci\'on Favorita Grocery Sales forecasting competition that shows daily unit sales of a set of items sold at different Favorita stores. The missing observations of this dataset are replaced by zeros. Covariates are also available with this dataset such as day of the week. 
    
    \item M5 Dataset \citep{makridakis_2020_m5}: A subset of time series from the M5 forecasting competition that shows daily unit sales of a set of items sold at different Walmart stores. To make the amount of time series comparable with the other datasets, we select the items that belong to the $HOBBIES\_2$ department which is one of the ten departments used in the competition.
    
    \item Tourism Monthly Dataset \citep{Athanasopoulos_2011_tourism}: Monthly dataset from the Tourism forecasting competition. The dataset was extracted from \citet{godahewa_2021_monash}.
    
    \item Tourism Quarterly Dataset \citep{Athanasopoulos_2011_tourism}: Quarterly dataset from the Tourism forecasting competition. The dataset was extracted from \citet{godahewa_2021_monash}.
    
    \item Chaotic Logistic Dataset: A simulated dataset used in \citet{hewamalage_2020_simulation} constructed by using the Chaotic Logistic Map \citep{may_1976_simple} Data Generation Process (DGP)

	\item Mackey-Glass Dataset: A simulated dataset used in \citet{hewamalage_2020_simulation} constructed by using the Mackey-Glass Equation \citep{mackey_1977_oscillation} DGP.
\end{itemize}

For our study, we have mostly used datasets where globally trained tree models are known to provide good forecasts. Note that four datasets: Rossmann, Kaggle Web Traffic, Favorita and M5 are from Kaggle competitions where tree-based models have won the first or second positions \citep{BOJER2020}. The two simulated datasets also contain non-linear time series where tree-based models have provided accurate forecasts compared to PR and neural network models based on the results published by \citet{hewamalage_2020_simulation}. To add some diversity into the pool of datasets, we consider the Tourism Monthly and Quarterly datasets from the Tourism forecasting competition where the traditional univariate forecasting models have provided better forecasts. There are other popular benchmarking datasets in the global forecasting field, most notably the M3 \citep{ref_5} and M4 \citep{ref_30} datasets. However, these datasets are very heterogeneous as they contain time series from many different use cases and require special preprocessing and/or ensembling techniques such as clustering and ensemble of specialists \citep{godahewa_2021_ensembling, ref_2, ref_1} to make global models competitive. As we deem such additional techniques to be out of the scope of our current paper, we do not consider these datasets for this study.

\subsection{Error metrics}
\label{sec:error_metrics}
We measure the performance of our models using the modified version of symmetric Mean Absolute Percentage Error \citep[msMAPE,][]{ref_14} and Mean Absolute Scaled Error \citep[MASE,][]{ref_36} which are commonly used error metrics in the time series forecasting field. Equations \ref{eqn:smape} and \ref{eqn:mase} respectively define the msMAPE and MASE error metrics. Here, $F_k$ are the forecasts, $Y_k$ are the actual values for the required forecast horizon, $M$ is the number of instances in the training set, $N$ is the number of data points in the test set and $S$ is the length of the seasonal cycle. In Equation \ref{eqn:smape}, the $\epsilon$ is set to its proposed default value of 0.1. 

\begin{equation}
\label{eqn:smape}
    msMAPE = \frac{100\%}{N}\sum_{k=1}^{N} \frac{\lvert F_{k} - Y_{k}\rvert}{(max(\lvert Y_{k}\rvert + \lvert F_{k}\rvert + \epsilon, 0.5 + \epsilon))/2} 
\end{equation}

\begin{equation}
\label{eqn:mase}
    MASE = \frac{\sum_{k=1}^{N} \lvert F_{k} - Y_{k}\rvert}{\frac{N}{M - S}\sum_{k=S+1}^{M} \lvert Y_{k} - Y_{k - S} \rvert}
\end{equation}

For the two simulated datasets, the length of the seasonal cycle is considered as one. Since all these error measures are defined for each time series, we calculate the mean and median values of them across a dataset to measure the model performance. Therefore, four error metrics are used to evaluate each model: mean msMAPE, median msMAPE, mean MASE and median MASE.

\subsection{Benchmarks and variants}
\label{sec:benchmarks}
We use seven global models: a PR model \citep{gelman_hill_2006}, Feed-Forward Neural Network \citep[FFNN,][]{Goodfellow-et-al-2016} and five tree-based models: a regression tree \citep{loh_2011_classification}, Cubist \citep{kuhn_2013_applied}, LightGBM \citep{guolin_lightgbm_2017}, XGBoost \citep{chen_xgboost_2016}, and CatBoost \citep{NEURIPS2018_14491b75}, as the main benchmarks of our study. These tree-based models are popular well-performing models in the forecasting domain. Regression trees have provided successful insights and foundations for machine learning approaches in the forecasting domain \citep{spiliotis_2022_trees}. Cubist is a well-known regression model tree that performs rule-based modelling \citep{quinlan1992learning}. Many top solutions of the M5 forecasting competition \citep{makridakis_2020_m5} including the winning method use globally trained LightGBM models. Furthermore, the winning methods of many recently held Kaggle forecasting competitions use tree-based models \citep{BOJER2020}. The FFORMA method \citep{ref_3} which placed second in the M4 forecasting competition \citep{ref_30} uses XGBoost as a meta-learner. CatBoost considers the order of data points during modelling which makes it more suitable for time series forecasting \citep{NEURIPS2018_14491b75}.

The R packages \verb|glmnet| \citep{ref_111}, \verb|nnet| \citep{ref_110}, \verb|rpart| \citep{therneau_2019_rpart}, \verb|Cubist| \citep{kuhn_2022_cubist}, \verb|lightgbm| \citep{ke_2020_lightgbm}, \verb|xgboost| \citep{chen_2020_xgboost} and \verb|catboost| \citep{NEURIPS2018_14491b75} are respectively used to implement the PR, FFNN, regression tree, Cubist, LightGBM, XGBoost and CatBoost models. The hyperparameters of FFNN, LightGBM and XGBoost are tuned using a grid search approach. For the FFNN, the number of nodes in the hidden layers is varied from 10 to 60 in steps of 10, and the learning rate decay is varied from 0.01 to 0.1 in steps of 0.01. For LightGBM, the minimum number of instances in a leaf node is varied from 50 to 200 in steps of 50, and the learning rate is varied from 0.01 to 0.1 in steps of 0.01. For XGBoost, the maximum tree depth is varied from 3 to 10 in steps of 1, and $\eta$ is varied from 0.1 to 0.5 in steps of 0.1. The regression tree, Cubist and CatBoost models are used with their default parameters that are assumed to provide good results. The PR model does not require any hyperparameter tuning.

We also consider two traditional univariate forecasting models: ETS \citep{ref_112} and ARIMA \citep{ref_113} that are commonly used in the forecasting space, as the benchmarks of our study. They are implemented using the R package, \verb|forecast| \citep{ref_22} under the default configurations.

The SETAR \citep{tong_1993_setar} and STAR \citep{Terasvirta_1994_star} models that are trained per series are also considered as benchmarks. They are implemented using the R package, \verb|tsDyn| \citep{narzo_tsdyn_2022} under the default configurations.

We further consider RF \citep{breiman_2001_random} as a benchmark to evaluate the performance of our SETAR-Forest algorithm. For that, the implementation of RF in the R package \verb|lightgbm| \citep{ke_2020_lightgbm} is used where the hyperparameters are tuned using a grid search approach. The minimum number of instances in a leaf node is varied from 50 to 200 in steps of 50, and the learning rate is varied from 0.01 to 0.1 in steps of 0.01. The bagging frequency, bagging fraction and feature fraction are respectively set to the same default values used in our SETAR-Forest algorithm: 10, 0.8 and 1 to facilitate the comparison between RF and the proposed SETAR-Forest model.

Three variants of the proposed SETAR-Tree model (Section \ref{sec: setar_tree_model}) are used as follows.

\begin{description}
\item \textbf{Tree.Lin.Test} Uses the significance of the linearity test as the stopping criterion of node splitting.

\item \textbf{Tree.Error.Red} Uses the error reduction percentage that can be gained by node splitting as the stopping criterion.

\item \textbf{Tree.Lin.Test.Error.Red} Uses both the significance of the linearity test and the error reduction percentage gained by node splitting as the stopping criteria.
\end{description}

Three variants of the proposed SETAR-Forest model (Section \ref{sec: setar_forest}) are used as follows.

\begin{description}
\item \textbf{Forest.Significance} The trees are randomised in terms of the significance level and the sequence significance divider considered in the linearity test.

\item \textbf{Forest.Error.Red} The trees are randomised in terms of the error threshold used to measure the error reduction percentage of node splitting.

\item \textbf{Forest.Significance.Error.Red} The trees are randomised in terms of the significance level and the sequence significance divider considered in the linearity test as well as the error threshold used to measure the error reduction percentage of node splitting.
\end{description}

\bigskip

The number of lagged values used in all GFMs including our SETAR-Tree and forest variants are determined using the heuristics suggested by \citet{ref_6} where the number of lags used with a dataset depends on its seasonal cycle length or forecast horizon. In particular, for daily and monthly datasets, we consider the heuristic, $seasonality \times 1.25$ to determine the number of lags. For daily datasets: Rossmann, Kaggle Web Traffic, Favorita and M5, this heuristic suggests 8.75 lags, where this value is rounded (to steps of 5) and 10 lags are finally considered with those datasets. For the Tourism Monthly dataset, 15 lags are considered as suggested by the above heuristic. For Chaotic Logistic, Mackey-Glass and Tourism Quarterly datasets, the seasonality-based heuristic suggests small values, so that for these datasets we use the horizon-based heuristic, and consider 10 lags which is equal to $horizon \times 1.25$ of those datasets. For the three datasets with external covariates: Rossmann, Kaggle Web Traffic and Favorita, the GFMs are separately executed with and without covariates where only the past lags are used when training without covariates, and the past lags and covariates are together used when training with covariates.

\subsection{Statistical testing of the results}
\label{sec: statistical_testing}

The non-parametric Friedman rank-sum test is used to assess the statistical significance of the results provided by different forecasting models across time series considering a significance level of $\alpha$ = 0.05 \citep{garcia_2010_advanced}. Based on the corresponding msMAPE errors, the methods are ranked on every series of the eight primary experimental datasets. In line with traditional forecasting modelling and evaluation \citep{KONING2005397}, we treat each series as a separate entity for statistical testing, so that the ranks per series are used as inputs for the statistical tests. This seems reasonable as the methods are univariate and -- once trained -- forecast each series in isolation. In this way, the statistical test has much more information instead of averaging the results over the datasets before applying the test. All primary experimental datasets have a comparable amount of series and thus, none of the datasets will dominate the evaluation and we obtain a representative result across all primary experimental datasets. The best method according to the average rank is chosen as the control method. To further characterise the statistical differences, Hochberg's post-hoc procedure is used \citep{garcia_2010_advanced}.

\section{Results}
\label{sec:results}
This section discusses the results in terms of main accuracy, statistical significance and computational performance.

\subsection{Main accuracy results}

This section first presents the results and a comparison of our SETAR-Tree and SETAR-Forest variants. The best SETAR-Tree and SETAR-Forest variants are then compared with the benchmarks.

\subsubsection{Comparison of SETAR-Tree and SETAR-Forest variants}
\label{sec: tree_forest_results}

\begin{table}
        \vspace{-1em}
        \caption{Results of SETAR-Tree and SETAR-Forest variants across all experimental datasets. The best performing variants in each group are italicized and the overall best performing variants are highlighted in boldface}
\begin{center}
        \resizebox{\textwidth}{!}{
		\begin{tabular}{rrrrrrrrrrrr}
			\toprule
			& \multicolumn{8}{c}{Without Covariates} & \multicolumn{3}{c}{With Covariates} \\
			\cmidrule(lr){2-9} \cmidrule(lr){10-12}
			& \ Ross- & Kag- & Favo-  & M5 & Tour & Tour   & Cha- & Mackey- & Ross- & Kag- & Favo-
      \\
      	& \ mann & gle & rita  & & (M) & (Q)   & otic & Glass & mann & gle & rita
      \\\hline
			\addlinespace
			\multicolumn{10}{l}{\bf Mean msMAPE} \\
			\addlinespace
Tree.Lin.Test & \textbf{39.05} & 71.04 & \textbf{\textit{83.01}} & \textbf{45.28} & 23.94 & 17.04 & \textbf{\textit{41.83}} & \textbf{\textit{0.00372}} & \textbf{11.53} & 66.64 & \textbf{94.77} \\ 
  Tree.Error.Red & 54.93 & \textbf{\textit{44.74}} & 85.04 & 53.92 & 22.62 & 19.04 & 49.36 & 0.00661 & 15.20 & 45.01 & 97.01 \\ 
  Tree.Lin.Test.Error.Red & 41.90 & \textbf{\textit{44.74}} & 85.04 & 53.92 & \textbf{\textit{21.52}} & \textbf{15.59} & 41.98 & \textbf{\textit{0.00372}} & 12.09 & \textbf{44.88} & 97.01 \\ 
  \hline
  Forest.Significance & 41.65 & 48.13 & 85.06 & \textbf{\textit{53.91}} & \textbf{21.17} & \textbf{15.59} & 41.30 & \textbf{0.00296} & 12.06 & \textbf{\textit{46.85}} & 96.67 \\ 
  Forest.Error.Red & 43.03 & 43.97 & 82.44 & 54.13 & 25.61 & 16.57 & 41.55 & 0.00307 & 12.07 & 47.94 & 95.32 \\ 
  Forest.Significance.Error.Red & \textbf{\textit{40.73}} & \textbf{43.80} & \textbf{82.36} & 54.13 & 22.16 & 15.97 & \textbf{41.14} & \textbf{0.00296} & \textbf{\textit{11.93}} & 47.83 & \textbf{\textit{95.28}} \\ 
            \bottomrule
			\addlinespace
			\multicolumn{10}{l}{\bf Median msMAPE} \\
			\addlinespace
Tree.Lin.Test & \textbf{36.86} & 69.72 & \textbf{81.88} & \textbf{41.48} & 19.61 & 13.44 & 36.37 & \textbf{0.00082} & \textbf{10.02} & 63.78 & \textbf{90.09} \\ 
  Tree.Error.Red & 42.90 & \textbf{\textit{41.52}} & 87.60 & 49.96 & 17.43 & 13.23 & 41.00 & 0.00118 & 11.64 & 40.95 & 91.33 \\ 
  Tree.Lin.Test.Error.Red & 41.07 & \textbf{\textit{41.52}} & 87.60 & 49.96 & \textbf{\textit{17.26}} & \textbf{\textit{12.59}} & \textbf{\textit{36.30}} & \textbf{0.00082} & 10.52 & \textbf{40.90} & 91.33 \\ 
  \hline
  Forest.Significance & 40.72 & 45.70 & 87.66 & \textbf{\textit{49.96}} & \textbf{16.96} & 12.62 & 37.41 & 0.00124 & 10.59 & \textbf{\textit{41.64}} & 91.12 \\ 
  Forest.Error.Red & 39.27 & 40.83 & \textbf{\textit{82.80}} & 50.08 & 18.40 & 12.70 & \textbf{36.07} & \textbf{0.00082} & \textbf{\textit{10.31}} & 43.20 & 90.85 \\ 
  Forest.Significance.Error.Red & \textbf{\textit{39.11}} & \textbf{40.81} & \textbf{\textit{82.80}} & 50.08 & 17.71 & \textbf{12.57} & 36.38 & 0.00124 & 10.38 & 43.17 & \textbf{\textit{90.81}} \\ 			
			\bottomrule
			\addlinespace
			\multicolumn{10}{l}{\bf Mean MASE} \\
			\addlinespace
Tree.Lin.Test & 0.546 & 2.605 & 0.835 & \textbf{1.330} & 1.726 & 1.698 & \textbf{\textit{0.682}} & \textbf{\textit{0.00641}} & \textbf{0.434} & 3.058 & 3.940 \\ 
  Tree.Error.Red & 0.676 & \textbf{\textit{0.960}} & \textbf{\textit{0.784}} & 1.442 & \textbf{\textit{1.572}} & 1.667 & 0.810 & 0.01150 & 0.571 & 1.428 & \textbf{\textit{1.015}} \\ 
  Tree.Lin.Test.Error.Red & \textbf{\textit{0.522}} & \textbf{\textit{0.960}} & \textbf{\textit{0.784}} & 1.442 & 1.575 & \textbf{\textit{1.559}} & 0.687 & \textbf{\textit{0.00641}} & 0.459 & \textbf{\textit{1.204}} & \textbf{\textit{1.015}} \\ 
  \hline
  Forest.Significance & 0.511 & 0.950 & 0.784 & \textbf{\textit{1.442}} & \textbf{1.546} & \textbf{1.550} & 0.670 & \textbf{0.00512} & 0.453 & \textbf{0.946} & 1.003 \\ 
  Forest.Error.Red & 0.506 & 0.895 & \textbf{0.757} & 1.445 & 1.627 & 1.589 & 0.676 & 0.00531 & 0.447 & 0.976 & 0.907 \\ 
  Forest.Significance.Error.Red & \textbf{0.456} & \textbf{0.872} & 0.777 & 1.445 & 1.601 & 1.593 & \textbf{0.667} & \textbf{0.00512} & \textbf{\textit{0.446}} & 0.966 & \textbf{0.906} \\
			\bottomrule
			\addlinespace
			\multicolumn{10}{l}{\bf Median MASE} \\
			\addlinespace
Tree.Lin.Test & \textbf{\textit{0.443}} & 1.872 & 0.737 & \textbf{1.037} & 1.532 & 1.428 & \textbf{\textit{0.650}} & \textbf{\textit{0.00135}} & \textbf{0.351} & 1.590 & \textbf{0.773} \\ 
  Tree.Error.Red & 0.518 & \textbf{\textit{0.776}} & \textbf{\textit{0.714}} & 1.138 & 1.409 & 1.365 & 0.742 & 0.00185 & 0.416 & \textbf{0.787} & 0.839 \\ 
  Tree.Lin.Test.Error.Red & 0.460 & \textbf{\textit{0.776}} & \textbf{\textit{0.714}} & 1.138 & \textbf{\textit{1.408}} & \textbf{\textit{1.320}} & 0.653 & \textbf{\textit{0.00135}} & 0.372 & \textbf{0.787} & 0.839 \\ 
  \hline
  Forest.Significance & 0.443 & 0.820 & 0.713 & \textbf{\textit{1.138}} & \textbf{1.401} & \textbf{1.291} & 0.657 & 0.00206 & 0.379 & \textbf{\textit{0.840}} & 0.828 \\ 
  Forest.Error.Red & 0.389 & \textbf{0.755} & \textbf{0.688} & 1.139 & 1.451 & 1.341 & \textbf{0.641} & \textbf{0.00128} & \textbf{\textit{0.370}} & 0.871 & \textbf{\textit{0.777}} \\ 
  Forest.Significance.Error.Red & \textbf{0.383} & \textbf{0.755} & \textbf{0.688} & 1.139 & 1.416 & 1.323 & 0.646 & 0.00206 & 0.371 & 0.869 & \textbf{\textit{0.777}} \\ 		
            \bottomrule
			\end{tabular}
		}	
		\label{tab:all_variant_results}
\end{center}
\end{table}

Table \ref{tab:all_variant_results} shows the results of the considered SETAR-Tree and SETAR-Forest variants across all experimental datasets for mean msMAPE, median msMAPE, mean MASE and median MASE. The tree variants and forest variants are separately grouped in Table \ref{tab:all_variant_results}. The results of the best performing variants in each group are italicized, and the
overall best performing variants across the datasets are highlighted in boldface. The preliminary experiments are also conducted with other stopping criteria such as using Akaike Information Criterion (AIC), however, the performance of the SETAR-Tree was not improved with them confirming the claims of prior similar studies \citep{rosa_str_2008}. Thus, the corresponding results are not included here.

Across the tree model variants, on mean msMAPE, the Tree.Lin.Test model variant shows the best performance across five primary datasets where the Tree.Lin.Test.Error.Red model variant shows the best performance across four primary datasets. However, on all other error metrics, Tree.Lin.Test.Error.Red shows the overall best performance across the majority of the experimental datasets from the tree model variants. Compared with the Tree.Lin.Test and Tree.Lin.Test.Error.Red model variants, the Tree.Error.Red model variant shows worse performance on all error metrics except across the Kaggle Web Traffic dataset. This shows that using the significance of the statistical linearity test individually or together with the error reduction percentage gained by node splitting are better options rather than using the error reduction percentage on its own as the stopping criterion of the SETAR-Tree to determine its maximum depth.

In the SETAR-Forest model variants, as the stopping criteria of each individual SETAR-Tree, we consider both the significance of the statistical linearity test and the error reduction percentage gained by node splitting which is the stopping criterion of our best SETAR-Tree variant, Tree.Lin.Test.Error.Red. Out of the three SETAR-Forest variants, Forest.Significance.Error.Red shows an overall better performance compared to the other two forest variants across all experimental datasets on all error metrics. This shows that when the individual SETAR-Trees of the forest are more diversified, the resultant forecasts are more accurate. We see that overall, the SETAR-Forest variants show the best performance across the majority of datasets compared to the SETAR-Tree variants on all error metrics. The forests properly address the data, model, and parameter uncertainties through bagging compared to the individual trees and that can be considered as the major reason for this phenomenon. 

Our SETAR-Tree and forest model variants also provide interesting results when trained with external covariates. As discussed in Section \ref{sec: covariates}, the categorical covariates of the Rossmann (status of the store, promotions, state holiday and school holiday flags), Kaggle Web Traffic (day of the week) and Favorita (day of the week) datasets are converted into the numerical format by applying a one-hot encoding mechanism before model training. Furthermore, with the Rossmann dataset, the daily number of customers to each store is considered as an external numerical covariate. In particular, the overall result of the Rossmann dataset has considerably improved across our proposed SETAR-Tree and forest variants after using external covariates on all error metrics. However, the results of the Kaggle Web Traffic and Favorita datasets do not show such improvements after using covariates across our proposed model variants.

\subsubsection{Comparison of proposed models with benchmarks}
\label{sec: benchmark_comparison}

Table \ref{tab:all_results} shows the results of the benchmarks and our best SETAR-Tree and SETAR-Forest variants, Tree.Lin.Test.Error.Red and Forest.Significance.Error.Red, across all experimental datasets for mean msMAPE, median msMAPE, mean MASE and median MASE.

\begin{table}
        \vspace{-1em}
        \caption{Results across all experimental
datasets. The best performing models in each group are italicized and the overall best performing models are highlighted in boldface}
\begin{center}
        \resizebox{\textwidth}{!}{
		\begin{tabular}{rrrrrrrrrrrr}
			\toprule
			& \multicolumn{8}{c}{Without Covariates} & \multicolumn{3}{c}{With Covariates} \\
			\cmidrule(lr){2-9} \cmidrule(lr){10-12}
			& \ Ross- & Kag- & Favo-  & M5 & Tour & Tour   & Cha- & Mackey- & Ross- & Kag- & Favo-
      \\
      	& \ mann & gle & rita  & & (M) & (Q)   & otic & Glass & mann & gle & rita
      \\\hline
			\addlinespace
			\multicolumn{10}{l}{\bf Mean msMAPE} \\
			\addlinespace
  ETS & \textbf{\textit{43.98}} & \textbf{\textit{46.24}} & \textbf{\textit{87.67}} & 78.22 & \textbf{19.02} & \textbf{15.07} & 50.33 & 1.02983 & - & - & - \\ 
  ARIMA & 45.34 & 47.96 & 87.82 & 77.81 & 19.73 & 16.58 & 48.71 & 11.12100 & - & - & - \\ 
  SETAR & 62.20 & 46.75 & 94.56 & \textbf{\textit{58.18}} & 31.30 & 36.14 & 52.93 & 0.04079 & - & - & - \\ 
  STAR & 72.89 & 46.82 & 96.30 & 95.01 & 32.58 & 34.08 & \textbf{\textit{44.82}} & \textbf{\textit{0.02094}} & - & - & - \\ 
  \hline
  PR & 64.45 & 111.48 & \textbf{\textit{85.04}} & 53.92 & 21.56 & 17.07 & 52.27 & \textbf{\textit{0.01949}} & 43.02 & 68.78 & 99.22 \\ 
  Cubist & \textbf{38.77} & 55.69 & 85.75 & 146.12 & \textbf{\textit{19.96}} & \textbf{\textit{16.02}} & 43.03 & 0.26995 & \textbf{\textit{13.07}} & 55.67 & \textbf{85.63} \\
  FFNN & 197.35 & 164.74 & 119.40 & 94.97 & 199.47 & 199.77 & 42.78 & 60.53347 & 197.35 & 164.74 & 115.14 \\ 
  Regression Tree & 55.48  &  61.88 & 101.21 & 65.94 & 64.34 & 115.02 & 44.72 & 3.42950 & 46.58  & 61.88 &  101.21 \\
  CatBoost & 49.39 & 49.66 & 90.73 & 57.33 & 23.75 & 25.37 & \textbf{\textit{42.09}} & 0.65735 & 39.91 & \textbf{\textit{47.97}} & 90.90 \\ 
  LightGBM & 56.16 & 55.63 & 96.83 & \textbf{32.60} & 22.18 & 19.72 & 42.53 & 0.56777 & 42.73 & 59.25 & 98.06 \\ 
  XGBoost & 48.29 & 69.73 & 86.93 & 54.99 & 23.48 & 18.84 & 44.40 & 0.45676 & 48.41 & 65.20 & 89.41 \\ 
  RF & 61.95 & \textbf{\textit{49.63}} & 103.02 & 104.49 & 32.55 & 27.13 & 42.62 & 2.85191 & 46.53 & 49.90 & 101.62 \\ 
   \hline
   Tree.Lin.Test.Error.Red & 41.90 & 44.74 & 85.04 & \textbf{\textit{53.92}} & \textbf{\textit{21.52}} & \textbf{\textit{15.59}} & 41.98 & 0.00372 & 12.09 & \textbf{44.88} & 97.01 \\ 
   Forest.Significance.Error.Red & \textbf{\textit{40.73}} & \textbf{43.80} & \textbf{82.36} & 54.13 & 22.16 & 15.97 & \textbf{41.14} & \textbf{0.00296} & \textbf{11.93} & 47.83 & \textbf{\textit{95.28}} \\ 
            \bottomrule
			\addlinespace
			\multicolumn{10}{l}{\bf Median msMAPE} \\
			\addlinespace
  ETS & \textbf{\textit{44.25}} & 41.72 & 86.59 & 72.62 & \textbf{17.16} & \textbf{\textit{12.89}} & 38.37 & 0.79343 & - & - & - \\ 
  ARIMA & 45.89 & 45.03 & \textbf{\textit{84.84}} & 73.08 & 18.00 & 13.13 & \textbf{\textit{37.26}} & 9.78094 & - & - & - \\ 
  SETAR & 57.45 & 41.55 & 88.75 & \textbf{\textit{46.39}} & 25.27 & 23.36 & 51.64 & 0.03111 & - & - & - \\ 
  STAR & 61.04 & \textbf{\textit{41.07}} & 89.65 & 76.06 & 23.66 & 17.34 & 39.53 & \textbf{\textit{0.01078}} & - & - & - \\ 
  \hline
  PR & 64.88 & 110.64 & 87.60 & 49.96 & 18.60 & 13.42 & 43.81 & \textbf{\textit{0.01255}} & 41.20 & 64.64 & 91.88 \\ 
  Cubist & \textbf{\textit{42.91}} & 53.62 & 87.79 & 148.76 & \textbf{\textit{17.24}} & \textbf{\textit{13.21}} & 39.93 & 0.21272 & \textbf{\textit{11.18}} & 53.62 & 87.69 \\ 
  FFNN & 197.29 & 168.20 & 114.39 & 91.12 & 199.84 & 199.94 & 37.18 & 62.12558 & 197.29 & 168.20 & 114.22 \\ 
  Regression Tree &  52.32 & 53.57 & 92.15 & 62.22 & 57.22 & 120.69 & 37.46 & 2.97342  & 44.85 & 53.57 & 92.15 \\
  CatBoost & 48.38 & 45.94 & 89.55 & 47.81 & 20.38 & 19.31 & \textbf{\textit{36.70}} & 0.48910 & 38.74 & \textbf{\textit{44.06}} & 89.77 \\ 
  LightGBM & 57.16 & 51.71 & 89.37 & \textbf{26.46} & 19.39 & 14.88 & 38.09 & 0.41013 & 40.92 & 55.78 & 89.23 \\ 
  XGBoost & 46.90 & 67.29 & \textbf{\textit{83.10}} & 47.26 & 21.05 & 16.19 & 37.39 & 0.34459 & 47.43 & 61.00 & \textbf{84.82} \\ 
  RF & 60.71 & \textbf{\textit{45.08}} & 95.50 & 102.75 & 27.16 & 20.74 & 37.13 & 2.16739 & 45.11 & 45.38 & 92.89 \\ 
   \hline
  Tree.Lin.Test.Error.Red & 41.07 & 41.52 & 87.60 & \textbf{\textit{49.96}} & \textbf{\textit{17.26}} & 12.59 & \textbf{36.30} & \textbf{0.00082} & 10.52 & \textbf{40.90} & 91.33 \\ 
  Forest.Significance.Error.Red & \textbf{39.11} & \textbf{40.81} & \textbf{82.80} & 50.08 & 17.71 & \textbf{12.57} & 36.38 & 0.00124 & \textbf{10.38} & 43.17 & \textbf{\textit{90.81}} \\ 
			\bottomrule
			\addlinespace
			\multicolumn{10}{l}{\bf Mean MASE} \\
			\addlinespace
 ETS & \textbf{\textit{0.584}} & 2.582 & 0.786 & 1.796 & \textbf{1.526} & \textbf{\textit{1.592}} & 0.770 & 1.61780 & -  & - & - \\ 
  ARIMA & 0.596 & 1.407 & \textbf{\textit{0.782}} & 1.807 & 1.589 & 1.782 & \textbf{\textit{0.743}} & 17.59567 & - & - & - \\ 
  SETAR & 1.260 & \textbf{\textit{0.897}} & 0.848 & \textbf{\textit{1.387}} & 2.225 & 2.591 & 0.851 & 0.06783 & - & - & - \\ 
  STAR & 1.380 & 0.902 & 0.857 & 1.540 & 2.110 & 1.910 & 0.788 & \textbf{\textit{0.03498}} & - & - & - \\ 
  \hline
  PR & 1.333 & 1.483 & 0.784 & 1.442 & 1.681 & \textbf{\textit{1.654}} & 0.802 & \textbf{\textit{0.03165}} & 0.638 & 4.040 & 1.106 \\ 
  Cubist & \textbf{\textit{0.602}} & \textbf{\textit{1.031}} & \textbf{\textit{0.780}} & 4.407 & \textbf{\textit{1.593}} & 1.693 & 0.696 & 0.43350 & 0.483 & \textbf{\textit{1.036}} & \textbf{0.779} \\ 
  FFNN & 3.167 & 1.901 & 0.977 & 2.259 & 9.747 & 12.568 & 0.699 & 77.63547 & 3.167 & 1.901 & 0.954 \\ 
  Regression Tree & 0.960  & 3.530 & 1.156 & 1.789 & 8.080 & 82.250 & 0.713 & 5.44867 & 0.667 & 3.530 & 1.156 \\
  CatBoost & 0.858 & 1.517 & 0.866 & 1.562 & 1.877 & 2.851 & \textbf{\textit{0.693}} & 1.06160 & \textbf{\textit{0.449}} & 1.584 & 0.860 \\ 
  LightGBM & 1.030 & 1.977 & 0.976 & \textbf{1.128} & 1.769 & 2.246 & 0.699 & 0.90972 & 0.539 & 2.660 & 1.060 \\ 
  XGBoost & 0.796 & 2.719 & 0.896 & 1.545 & 1.924 & 2.094 & 0.722 & 0.75285 & 0.710 & 2.753 & 0.850 \\ 
   RF & 1.167 & 1.901 & 1.325 & 2.522 & 2.596 & 3.431 & 0.705 & 4.46711 & 0.663 & 1.913 & 1.175 \\ 
    \hline
    Tree.Lin.Test.Error.Red & 0.522 & 0.960 & 0.784 & \textbf{\textit{1.442}} & \textbf{\textit{1.575}} & \textbf{1.559} & 0.687 & 0.00641 & 0.459 & 1.204 & 1.015 \\ 
  Forest.Significance.Error.Red & \textbf{0.456} & \textbf{0.872} & \textbf{0.777} & 1.445 & 1.601 & 1.593 & \textbf{0.667} & \textbf{0.00512} & \textbf{0.446} & \textbf{0.966} & \textbf{\textit{0.906}} \\
			\bottomrule
			\addlinespace
			\multicolumn{10}{l}{\bf Median MASE} \\
			\addlinespace
  ETS & \textbf{\textit{0.552}} & 0.800 & \textbf{0.683} & 1.454 & \textbf{1.276} & \textbf{1.275} & 0.697 & 1.32630 & - & - & - \\ 
  ARIMA & 0.585 & 0.808 & 0.684 & 1.403 & 1.337 & 1.388 & \textbf{\textit{0.686}} & 16.15742 & - & - & - \\ 
  SETAR & 1.144 & 0.779 & 0.765 & \textbf{\textit{1.087}} & 1.840 & 2.316 & 0.844 & 0.04834 & - & - & - \\ 
  STAR & 1.250 & \textbf{\textit{0.761}} & 0.771 & 1.482 & 1.689 & 1.664 & 0.706 & \textbf{\textit{0.01861}} & - & - & - \\ 
  \hline
  PR & 1.291 & 1.368 & 0.714 & 1.138 & 1.502 & 1.442 & 0.778 & \textbf{\textit{0.02055}} & 0.480 & 1.531 & 0.886 \\ 
  Cubist & \textbf{\textit{0.562}} & 0.893 & \textbf{\textit{0.697}} & 3.399 & \textbf{\textit{1.406}} & \textbf{\textit{1.432}} & 0.705 & 0.35068 & 0.391 & 0.894 & \textbf{0.705} \\ 
  FFNN & 2.981 & 1.705 & 0.808 & 1.848 & 9.105 & 11.614 & 0.688 & 78.37429 & 2.981 & 1.705 & 0.798 \\ 
  Regression Tree & 0.807  & 1.108 & 0.873 & 1.484 & 4.440 & 19.586 & 0.685 & 4.71614 & 0.557 & 1.108 & 0.873 \\
  CatBoost & 0.822 & 0.896 & 0.742 & 1.235 & 1.668 & 2.000 & \textbf{\textit{0.664}} & 0.75808 & \textbf{0.371} & \textbf{\textit{0.854}} & 0.737 \\ 
  LightGBM & 1.017 & 1.055 & 0.793 & \textbf{0.842} & 1.569 & 1.711 & 0.674 & 0.70062 & 0.431 & 1.238 & 0.833 \\ 
  XGBoost & 0.735 & 1.760 & 0.761 & 1.220 & 1.777 & 1.737 & 0.683 & 0.56261 & 0.659 & 1.479 & 0.734 \\ 
   RF & 1.095 & \textbf{\textit{0.865}} & 0.956 & 2.035 & 2.080 & 2.174 & 0.682 & 3.86985 & 0.554 & 0.865 & 0.889 \\ 
    \hline
 	  Tree.Lin.Test.Error.Red & 0.460 & 0.776 & 0.714 & \textbf{\textit{1.138}} & \textbf{\textit{1.408}} & \textbf{\textit{1.320}} & 0.653 & \textbf{0.00135} & 0.372 & \textbf{0.787} & 0.839 \\ 
    Forest.Significance.Error.Red & \textbf{0.383} & \textbf{0.755} & \textbf{\textit{0.688}} & 1.139 & 1.416 & 1.323 & \textbf{0.646} & 0.00206 & \textbf{0.371} & 0.869 & \textbf{\textit{0.777}} \\ 	
            \bottomrule
			\end{tabular}
		}	
		\label{tab:all_results}
\end{center}
\end{table}

The models in Table \ref{tab:all_results} are grouped based
on the sub-experiments. The results of the best performing models in each group are italicized, and the overall best performing models across the datasets are highlighted in boldface. The first group contains the traditional univariate forecasting models, ETS, ARIMA, SETAR and STAR, where these models are only executed with our eight primary experimental datasets without using external covariates. The second group contains benchmark GFMs including PR, FFNN, state-of-the-art tree-based models and RF. The last group contains our best SETAR-Tree and SETAR-Forest variants that are discussed in Section \ref{sec: tree_forest_results}. The GFMs are executed with and without covariates for three datasets: Rossmann, Kaggle Web Traffic and Favorita, separately.

In the first group, ETS shows an overall better performance compared to ARIMA, SETAR and STAR across all error metrics. ETS shows the best performance across the Tourism Monthly dataset on all error metrics, Tourism Quarterly dataset on mean msMAPE and median MASE, and the Favorita dataset on median MASE. However, GFMs including our proposed models show a better performance than these traditional univariate forecasting models across all experimental datasets on all error metrics except for these seven cases.

In the second group, overall, PR and Cubist show a better performance compared to the other benchmark GFMs across all error metrics. The Cubist and CatBoost models show a better performance when training with external covariates.

The third group shows the results of our best SETAR-Tree variant, Tree.Lin.Test.Error.Red, and the best SETAR-Forest variant, Forest.Significance.Error.Red. Our proposed models show an overall better performance than other benchmark GFMs listed in the second group. In particular, our proposed models outperform the state-of-the-art tree-based and forest-based forecasting algorithms such as the regression tree, Cubist, LightGBM, XGBoost, CatBoost and RF across all experimental datasets on all error metrics except for twelve cases: Rossmann on mean msMAPE (Cubist), M5 on all error metrics (LightGBM), M5 on median msMAPE (CatBoost and XGBoost), Tourism Monthly on mean msMAPE, median msMAPE and median MASE (Cubist), Favorita with covariates on median msMAPE (LightGBM) and on all error metrics (Cubist, Catboost and XGBoost). This further shows training a global AR model in the leaf nodes leads to more accurate forecasts rather than simply considering the average of the training instances in the leaf nodes. We also see the results of PR and Tree.Lin.Test.Error.Red are the same across the Favorita dataset without covariates and the M5 dataset on all error metrics. This means the corresponding SETAR-Tree have only one node which is the parent node where the parent node has not been further split based on the considered stopping criterion. The SETAR-Tree and the PR model have overall provided better results across the above two datasets compared to most of the other benchmark GFMs. It shows limiting the tree to a single node is a worthwhile decision in these cases. We see that overall, Forest.Significance.Error.Red shows the best performance across the majority of datasets compared to all other considered benchmarks listed in the first and second groups as well as Tree.Lin.Test.Error.Red.

Overall, our proposed models show better performance when training with covariates compared to the benchmarks including PR, FFNN and the state-of-the-art tree-based algorithms. As explained in Section \ref{sec: tree_forest_results}, the overall result of the Rossmann dataset has considerably improved across our proposed models after using external covariates on all error metrics while considerably outperforming all benchmarks. The results of the Kaggle Web Traffic and Favorita datasets do not show such improvements after using covariates across our proposed models as well as all other considered benchmarks. However, across the Kaggle Web Traffic dataset, Tree.Lin.Test.Error.Red shows the overall best performance compared to the other benchmarks when training with covariates.

\subsubsection{Analysis of SETAR-Forest size}
\label{sec: forest_analysis}

\begin{table}
\begin{center}
 \caption{Results of Forest.Significance.Error.Red for Chaotic Logistic dataset across four different sizes. The best performing models are highlighted in boldface}
		\resizebox{\textwidth}{!}{
		\begin{tabular}{rrrrr}
			\toprule
		No: of Trees  & Mean msMAPE &  Median msMAPE & Mean MASE & Median MASE
      \\\hline
			\addlinespace
   5 & 41.65 & 36.57 &  0.676 & \textbf{0.644} \\ 
   10 & \textbf{41.14} & 36.38 & \textbf{0.667} & 0.646 \\ 
   20 & 41.34 & 35.65 &  0.670 & \textbf{0.644} \\ 
   50 & 41.25 & \textbf{35.60} & 0.668 & 0.647 \\ 
 \bottomrule
 \end{tabular}
         }
\label{tab:forest_size}
\end{center}
\end{table}

We analyse the effect of the size of the SETAR-Forest, i.e., the number of SETAR-Trees in the forest, for the final forecasting accuracy. Table \ref{tab:forest_size} shows the results of our best SETAR-Forest variant, Forest.Significance.Error.Red, for the Chaotic Logistic dataset across four different sizes: 5, 10, 20 and 50 for mean msMAPE, median msMAPE, mean MASE and median MASE. 

As expected, based on the results, overall, the performance of the SETAR-Forest gets better when the number of SETAR-Trees is increased. The SETAR-Forest containing 10 SETAR-Trees shows the best performance on mean msMAPE and mean MASE and the second best performance on median MASE. Adding more trees does not lead to considerable accuracy gains. Thus, we deem ten SETAR-Trees sufficient for a SETAR-Forest to obtain accurate forecasts in our experiments.

\subsection{Statistical testing results}

\begin{table}
\begin{center}
\begin{minipage}{174pt}
\caption{Results of statistical testing} \label{tab:statistical_testing}
\begin{tabular}{lc}
  \hline
  Model & $p_{Hoch}$ \\ 
  \hline
   Forest.Significance.Error.Red &  -   \\
   \hline
   Tree.Lin.Test.Error.Red & $0.003$     \\
   LightGBM &    $< 10^{-30}$ \\
   ARIMA &     $< 10^{-30}$ \\
   ETS &    $< 10^{-30}$ \\
   CatBoost &    $< 10^{-30}$ \\
   SETAR & $< 10^{-30}$ \\
   XGBoost &    $< 10^{-30}$ \\
   STAR & $< 10^{-30}$  \\
   Cubist & $< 10^{-30}$  \\
   PR &  $< 10^{-30}$    \\
   Regression Tree &    $< 10^{-30}$ \\
   RF & $< 10^{-30}$    \\
   FFNN &   $< 10^{-30}$  \\
  \hline
\end{tabular}
\end{minipage}
\end{center}
\end{table}

Table \ref{tab:statistical_testing} shows the results of the statistical testing evaluation, namely the adjusted $p$-values calculated from the Friedman test with Hochberg's post-hoc procedure considering a significance level of $\alpha = 0.05$ \citep{garcia_2010_advanced}. We consider the msMAPE of each series provided by the benchmarks and our best SETAR-Tree and SETAR-Forest variants, Tree.Lin.Test.Error.Red and Forest.Significance.Error.Red, for the eight primary experimental datasets, namely the Rossmann, Kaggle Web Traffic, Favorita, M5, Tourism Monthly, Tourism Quarterly, Chaotic Logistic, and Mackey-Glass datasets. The datasets are not considered with the external covariates during statistical testing as ETS, ARIMA, SETAR and STAR models are only executed without covariates. 

The overall $p$-value of the Friedman rank sum test is less than $10^{-30}$ which is highly significant. Forest.Significance.Error.Red performs the best on ranking over msMAPE per each series of the eight primary experimental datasets and thus, it is used as the control method as mentioned in the first row. A horizontal line is used to separate the models that perform significantly worse than Forest.Significance.Error.Red. All benchmarks and Tree.Lin.Test.Error.Red are significantly worse than the control method as they report $p_{Hoch}$ values less than $\alpha$. 

\subsection{Computational performance}

\begin{table}
\begin{center}
 \caption{Computational times (in minutes) of all models across all experimental datasets}
		\resizebox{\textwidth}{!}{
		\begin{tabular}{rrrrrrrrrrrr}
			\toprule
		& \multicolumn{8}{c}{Without Covariates} & \multicolumn{3}{c}{With Covariates} \\
			\cmidrule(lr){2-9} \cmidrule(lr){10-12}
			& \ Ross- & Kag- & Favo-  & M5 & Tour & Tour   & Cha- & Mackey- & Ross- & Kag- & Favo-
      \\
      	& \ mann & gle & rita  & & (M) & (Q)   & otic & Glass & mann & gle & rita
      \\\hline
			\addlinespace
  ETS & 7.52 & 5.59 & 8.08 &  14.43  & 5.00  & 1.06 & 0.07 & 0.22 & - & - & - \\ 
  ARIMA & 163.80 & 11.16 & 123.00 &  25.41  & 47.00 & 6.78 & 0.18 & 0.30 & - & - & - \\ 
  SETAR & 7.55 & 0.90 & 2.01 & 5.17 &  0.59  &  0.27 & 0.38 & 1.29 & - & - & - \\ 
  STAR & 99.68 & 70.33 & 133.87 &  140.29  & 10.04 & 1.86 & 4.81 & 5.11 & - & - & - \\ 
  PR & 0.55 & 0.63 & 0.65 &  1.58  & 0.03 & 0.02 & 0.02 & 0.02 & 0.90 & 0.25 & 0.55 \\ 
  Cubist & 1.51 & 7.40 & 2.55 &  7.44  &  0.52 &  0.27 & 0.10 & 0.06 & 1.86 & 2.16 & 2.73  \\ 
  FFNN & 7.57 & 18.88 & 223.80 &  24.01  & 0.74  & 0.25 & 25.00 & 4.61 & 10.16 & 24.22 & 352.20 \\ 
  Regression Tree & 0.98 & 0.48 & 1.31 & 1.84   &  0.02 & 0.02 & 0.03 & 0.02 & 1.22 & 0.37 & 1.60 \\ 
  CatBoost & 2.13 & 2.17 & 1.12 &  2.07  & 0.24  & 0.13 & 0.15 & 0.43 & 6.07 & 1.20 & 2.99 \\ 
  LightGBM & 3.24 & 12.73 & 5.25 &  8.91  &  0.48 & 5.03 & 3.84 & 6.27 & 48.02 & 5.44 & 12.22 \\ 
  XGBoost & 81.60 & 45.00 & 231.60 &  70.26 &  5.23  &  3.75 & 7.37 & 4.72 & 341.40 & 21.93 & 59.53 \\ 
  RF & 1.49 & 58.30 & 2.76 &  4.71  & 6.89 & 0.10 & 0.28 & 0.12 & 10.43 & 6.44 & 14.29 \\ 
  Tree.Lin.Test.Error.Red & 13.85 & 5.25 & 0.24 &  0.53  & 0.34 & 0.12 & 0.09 & 19.25 & 38.78 & 14.52 & 3.22 \\ 
  Forest.Significance.Error.Red & 124.16 & 68.22 & 35.56 &  5.53  & 7.14  & 1.79 & 0.77 & 103.82 & 310.60 & 175.88 & 300.15 \\
 \bottomrule
 \end{tabular}
         }
\label{tab:computational_times}
\end{center}
\end{table}

\begin{table}
\begin{center}
 \caption{Times (in seconds) required by the proposed models to predict a data point (given the fully trained models). The prediction time of Forest.Significance.Error.Red is ten times higher than the prediction time of Tree.Lin.Test.Error.Red as we consider ten SETAR-Trees for a SETAR-Forest.}
		\resizebox{\textwidth}{!}{
		\begin{tabular}{rrrrrrrrrrrr}
			\toprule
		& \multicolumn{8}{c}{Without Covariates} & \multicolumn{3}{c}{With Covariates} \\
			\cmidrule(lr){2-9} \cmidrule(lr){10-12}
			& \ Ross- & Kag- & Favo-  & M5 & Tour & Tour   & Cha- & Mackey- & Ross- & Kag- & Favo-
      \\
      	& \ mann & gle & rita  & & (M) & (Q)   & otic & Glass & mann & gle & rita
      \\\hline
			\addlinespace
  Tree.Lin.Test.Error.Red & 1.20   &  0.65  &  0.01  &  0.01   &  0.29  &  0.32  &  0.06  &  0.16  & 1.52 & 0.96   &   0.87 \\ 
  Forest.Significance.Error.Red &  12.00  & 6.50   & 0.10    & 0.10    &    2.90 & 3.20   &  0.60  &    1.60 & 15.20 & 9.60   & 8.70   \\ 
 \bottomrule
 \end{tabular}
         }
\label{tab:throughput}
\end{center}
\end{table}
To compare the computational performance of the benchmarks and our proposed models, all experiments are executed in a controlled environment, namely an Intel(R) Core(TM) i7 processor (2.6GHz) and 32GB of main memory.

\begin{sloppypar}
Table \ref{tab:computational_times} shows the computational times (in minutes) of all benchmarks and our proposed tree and forest models across all experimental datasets. The reported computational times include the model training, hyperparameter tuning and forecast computation times. 
\end{sloppypar}

From Table \ref{tab:computational_times}, we can see that the simple models such as PR and regression tree show the lowest computational times. Cubist, CatBoost and LightGBM models generally show lower computational times compared to XGBoost. Our proposed SETAR-Tree model, Tree.Lin.Test.Error.Red, overall shows lower computational times than Cubist, LightGBM and XGBoost across all primary datasets except Rossmann and Mackey-Glass. The most accurate model which is our proposed SETAR-Forest model, Forest.Significance.Error.Red, generally shows higher computational times than the benchmarks as it executes multiple SETAR-Trees and combines their forecasts to produce the final forecasts where a single tree internally does a lot of automatic parameter tuning. However, our SETAR-Forest shows lower computational times across the M5, Tourism Quarterly and Chaotic Logistic datasets than the LightGBM and XGBoost models. RF shows lower computational times than the SETAR-Forest model, however, as our SETAR-Forest is more accurate than the RF, the higher computational times seem justified.  

Table \ref{tab:throughput} shows the times (in seconds) required by the proposed models, once trained, to predict a data point. Across the majority of datasets, the proposed SETAR-Tree model takes less than a second for prediction. The proposed SETAR-Forest model takes a few seconds for the same prediction, given the trained models. Thus, we deem our models computationally feasible. 
 
\bigskip

Based on our results, we recommend to use our best SETAR-Tree variant, Tree.Lin.Test.Error.Red, and the best SETAR-Forest variant, Forest.Significance.Error.Red, over the state-of-the-art tree-based algorithms.

\section{Conclusions and future research}
\label{sec:concluson}
Globally trained tree-based models have recently become popular in the forecasting field due to their contribution to most top-performing methods in the M5 forecasting competition. However, the models used are general-purpose models not specific to the forecasting task, and they only consider the average of the training instances at a leaf node as their prediction. 

In this paper, we have proposed a new forecasting-specific globally trained tree model and a forest model, the SETAR-Tree and SETAR-Forest, which use time-series-specific splitting and stopping procedures. The SETAR-Tree uses the underlying concept of SETAR models during node splitting. In contrast to the state-of-the-art tree-based algorithms, our proposed tree model trains a global PR model in each leaf node allowing the models to learn the cross-series information. The SETAR-Tree internally controls the tree depth by conducting a statistical linearity test and measuring the error reduction percentage at each node split while making the requirement of external hyperparameter tuning to a minimal state. The SETAR-Forest model uses a collection of SETAR-Trees during forecasting where the trees are made diverse by randomising the significance of the statistical linearity test and the threshold of the error reduction percentage used during node splitting. The proposed tree and forest models can also be trained with external covariates. Across eight experimental datasets, we have shown that our proposed tree and forest models can significantly outperform a set of state-of-the-art tree-based GFMs and traditional univariate forecasting models. As our proposed models show a better forecasting accuracy, with less parameter tuning with a competitive computational performance compared to the benchmarks, we recommend SETAR-Tree and SETAR-Forest as strong tree-based models for global time series forecasting.

From our experiments, we conclude that training a global model in the leaf nodes often leads to more accurate results in tree-based algorithms rather than considering the average of the training instances in the leaf nodes in the forecasting context. We further conclude that considering the significance of the statistical linearity test and the error reduction percentage gained by node splitting together is a good option in automatically determining the maximum tree depth. The SETAR-Forest shows the best performance across the majority of the datasets as it adequately addresses the data, model, and parameter uncertainties compared to the individual tree models. The results get further improved when the individual trees in the forest are more diversified.

The success of this approach encourages as future work to extend the proposed tree models to work with multiple regimes, to work with boosting techniques and to provide probabilistic forecasts. It will be also interesting to study whether incorporating tree pruning can further increase the forecasting accuracy of SETAR-Trees. Analysing the effects of training a GFM in the leaf nodes of the state-of-the-art tree-based algorithms such as LightGBM and CatBoost is also worth to study.

\section*{Declarations}

\begin{itemize}
\item \textbf{Funding -} This work was supported by the Australian Research Council under grant DE190100045, a Facebook Statistics for Improving Insights and Decisions research award and Monash University Graduate Research funding.

\item \textbf{Competing interests -} The authors have no relevant financial or non-financial interests to disclose.

\item \textbf{Ethics approval -} Not applicable

\item \textbf{Consent to participate -} Not applicable

\item \textbf{Consent for publication -} The paper is the authors' original work and has not been published nor has been submitted simultaneously elsewhere. All the authors have checked the manuscript and have agreed to the submission.

\item \textbf{Availability of data and materials -} All experimental datasets are publicly available at \url{https://github.com/rakshitha123/SETAR_Trees/tree/master/datasets}. All experimental results are mentioned and explained in the manuscript.

\item \textbf{Code availability -} All implementations of our work are publicly available at \url{https://github.com/rakshitha123/SETAR_Trees}.

\item \textbf{Authors' contributions - Conceptualisation:} Rakshitha Godahewa, Geoffrey I. Webb, Christoph Bergmeir and Daniel Schmidt; \textbf{Methodology:} Rakshitha Godahewa and Christoph Bergmeir; \textbf{Software:} Rakshitha Godahewa and Daniel Schmidt; \textbf{Formal analysis and investigation:} Rakshitha Godahewa and Christoph Bergmeir; \textbf{Writing - original draft preparation:} Rakshitha Godahewa; \textbf{Writing - review and editing:} Rakshitha Godahewa, Geoffrey I. Webb, Christoph Bergmeir and Daniel Schmidt; \textbf{Supervision:} Geoffrey I. Webb, Christoph Bergmeir and Daniel Schmidt.
\end{itemize}

\bibliography{sample}

\end{document}